\documentclass[letterpaper, 10 pt, conference]{ieeeconf}  

\IEEEoverridecommandlockouts                              
\usepackage[utf8]{inputenc}
\usepackage[T1]{fontenc}
\usepackage{svg}
\usepackage{lscape}
\usepackage{colortbl}
\usepackage{soul}
\title{\LARGE \bf
Food Recommendation as Language Processing (F-RLP): \\ A Personalized and Contextual Paradigm 
}

\author{Ali Rostami$^{1}$, Ramesh Jain$^{1}$, and Amir M. Rahmani$^{1}$ 
\thanks{$^{1}$Ali Rostami, Ramesh Jain and Amir M. Rahmani are with the Department of Computer
Science and Institute for Future Health (IFH), University of California,
Irvine, California, USA}%
}

\begin{document}

\maketitle
\thispagestyle{empty}
\pagestyle{empty}

\begin{abstract}

State-of-the-art rule-based and classification-based food recommendation systems face significant challenges in becoming practical and useful. This difficulty arises primarily because most machine learning models struggle with problems characterized by an almost infinite number of classes and a limited number of samples within an unbalanced dataset. Conversely, the emergence of Large Language Models (LLMs) as recommendation engines offers a promising avenue. However, a general-purpose Recommendation as Language Processing (RLP) approach lacks the critical components necessary for effective food recommendations. To address this gap, we introduce Food Recommendation as Language Processing (F-RLP), a novel framework that offers a food-specific, tailored infrastructure. F-RLP leverages the capabilities of LLMs to maximize their potential, thereby paving the way for more accurate, personalized food recommendations.

\end{abstract}


\section{INTRODUCTION}

In an era of ubiquitous digital assistants and tailored experiences, personalized food recommendations have emerged as a potent force shaping our culinary journey. Beyond mere convenience, these systems hold immense potential to improve dietary choices \cite{tran2021recommender}, address nutritional deficiencies \cite{chen2021personalized}, and even combat chronic diseases \cite{agapito2018dietos}.
Personalization in food recommendations transcends mere taste preferences. By incorporating underlying food-specific data of ingredients and recipes, cultural factors, health data, and real-time context, these systems can foster healthier and more fulfilling culinary experiences \cite{Rostami21PPFM}. Studies show that personalized recommendations can encourage healthier food choices, increase dietary adherence, and improve overall user satisfaction \cite{Rostami20PFM}. This intersection of personal well-being and culinary delight underscores the profound societal and individual benefits of effective food recommendation systems.


Food recommendation systems encompass a diverse array of approaches, from content-based filtering relying on user-rated recipes to collaborative filtering leveraging shared preferences \cite{ornab2017empirical}. Recent advancements have demonstrated the integration of hybrid and recommendation fusion techniques to cater to the multi-dimensionality of food choices \cite{Rostami20PFM}. Moreover, the rise of LLMs has sparked immense interest in their potential to revolutionize food recommendations \cite{Aljbawi2020HealthawareFP}.
LLMs hold the promise of understanding the linguistic nuances of food descriptions, user preferences, and context, paving the way for highly personalized and contextually aware recommendations \cite{geng2022recommendation}. However, existing LLM-based recommendation systems often lack a holistic approach, struggling to seamlessly integrate the diverse components crucial for effective food recommendations \cite{Podszun2023}.


Despite significant advancements in food recommendation systems, there remains a pronounced gap between user expectations and the performance of existing technologies. These systems frequently fall short in offering personalized suggestions, failing to account for essential factors such as dietary restrictions, cultural preferences, and the real-time context of users. Current methods that leverage LLM-based algorithms face challenges in accurately interpreting the nuanced language of food, which is critical for generating meaningful and customized recommendations. This limitation undermines their ability to truly personalize the user experience and maximize the relevance of their suggestions \cite{Aljbawi2020HealthawareFP}.
This disconnect between potential and reality underscores the need for a paradigm shift in food recommendation systems, one that leverages the power of LLMs while addressing their limitations. Geng \textit{et al.} \cite{geng2022recommendation}  introduce a promising methodology to harness LLMs for recommendation purposes, termed Recommendation as Language Processing (RLP). While this represents a significant advancement in utilizing LLMs for recommendations, the demand for a food-centric approach to enhance food-specific recommendations remains pronounced. A system that successfully bridges this gap has the potential to transform our interactions with food, promoting healthier, more satisfying, and culturally informed culinary experiences.


This paper proposes Food Recommendation as Language Processing, F-RLP, as a novel framework consisting of two primary elements: the first component is designed to aggregate and process food recommendation-specific data, employing a series of mathematical and logical operations to distill this information into a final vector. This vector is then inputted into a LLM. The second component, an LLM-based recommendation system, is adept at interpreting this contextual and personalized vector, utilizing it to generate accurate and tailored food recommendations.
Our contribution lies in the following key aspects:
\begin{itemize}
    \item Comprehensive food recommendation framework: Our hybrid framework integrates various personal and contextual data, with a specific focus on numerical data, which traditionally poses a challenge for LLMs. 
    \item Specialized LLM training: We propose a novel training regimen, centered around the use of enhanced counterfactual data allowing expert insights affect fine-tuning. 
    \item Novel context injection: F-RLP  introduces a context injection technique to the LLMs, streamlining the process of complex mathematical computations. This is achieved by supplying the LLM with a context-oriented list of options in conjunction with the user's query. 
\end{itemize}
\begin{figure}
    \centering
    \includegraphics[width=0.47\textwidth]{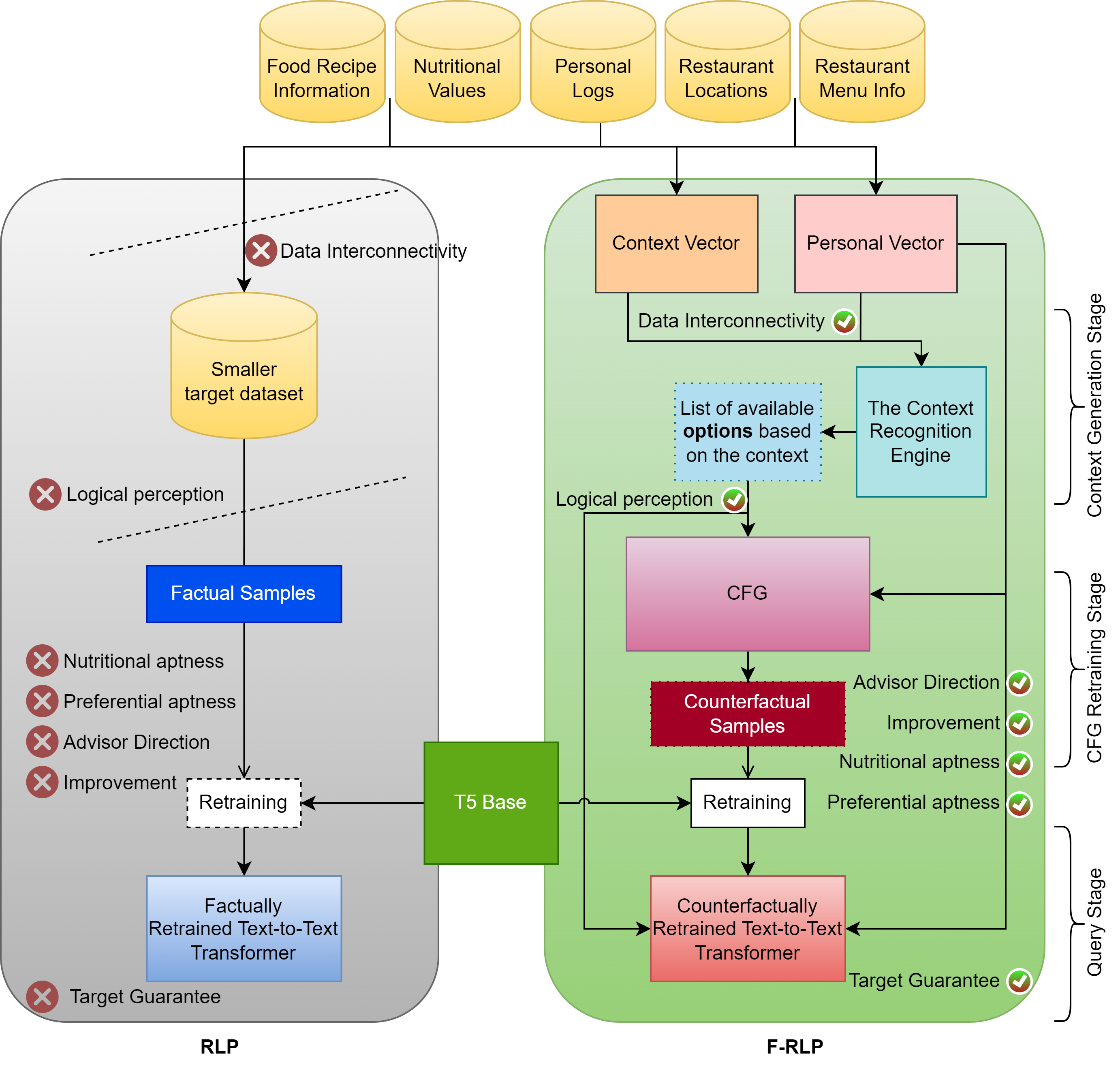}
    \caption{RLP vs F-RLP}
    \label{fig:arch}
\end{figure}


\section{F-RLP Paradigm and Model}
The F-RLP paradigm harnesses the power of LLMs to infuse food-specific personalizing and context into the broader RLP framework. In this paper, we adopt the T5 model introduced by \cite{t5} as our base RLP. F-RLP preparation includes two main buildup steps and a utility step:

\begin{enumerate}
    \item Context Generation Stage: This stage involves supplying the LLM with a curated list of contextualized options within each query, ensuring that the responses are practical and appropriately aligned with the user's context and location.
    \item Counterfactual Generation (CFG) Retraining Stage:  In this stage, the foundational LLM is retrained using counterfactual data, tailored to be both personalized and improvement-oriented. This retraining process, facilitated by the CFG component's design, aims to imbue the model with the capability to offer added value to the user.
    \item Query Stage: Occurring after the LLM has been fully prepared, this stage is when the user seeks recommendations. Our framework enriches this process by incorporating specific auxiliary data into the query to refine the recommendation process, details of which will be explored subsequently.
\end{enumerate}

As shown in Figure 1, F-RLP consists of three stages, the first two are training and buildup steps and the last stage is a utility step. 
In comparing F-RLP with RLP, F-RLP distinguishes itself primarily during the Context Generation Stage by utilizing both personal and contextual vectors to curate a list of options. This list is subsequently utilized in the CFG Retraining Stage, and then tailored to individual queries during the Query Stage. A significant innovation in the CFG Retraining Stage under F-RLP involves the generation of counterfactual data, which enriches the training dataset with nutritional and preference-based enhancements—a feature RLP does not incorporate. Furthermore, each query in the Query Stage is crafted with a personal vector and a set of contextual options, ensuring targeted selection from the curated list. This mechanism represents another key advantage of F-RLP over RLP. The query process leverages a Counterfactually Retrained Text-to-Text Transformer, illustrating F-RLP's advanced approach to food recommendation. In the following, we will delve deeper into these stages. 

\subsection{Context Generation Stage}
In this stage, we convert the user's food log along with other personal data, such as location and sensory information gathered from wearables and smartphones, into a Context Vector and a Personal Vector. We employ a simplified model of the context and personal vector representation as introduced by \cite{Rostami21PPFM}. These vectors are then processed by the Context Recognition Engine (CRE) to generate a list of contextual food choice options available to the user, taking into account their current location and context.

It is crucial to highlight the significance of the CRE's existence and its role within the overall architecture. By considering location and context, the CRE plays a fundamental part in curating the contextual list of options. The design of the CRE draws upon our prior work. For further information on designing a CRE, readers are directed to \cite{Rostami21PPFM} and \cite{Rostami2022WFA}. 
Methods including query lookup on geospatial restaurant datasets, as introduced by \cite{Rostami21WFA} and \cite{Rostami2022WFA}, can be utilized to fulfill the implementation requirements of this component. In our experiment, we deploy a simplified version of the Context Recognition Engine (CRE), which selects twenty food choices at random from the \cite{marin2021recipe1m+} dataset. This dataset not only provides the nutritional content of each food choice but also details its ingredients. We employ these simplified CRE samples to demonstrate the effectiveness of the F-RLP methodology in practice.

\subsection{Counterfactual Generation (CFG) Retraining Stage}
Figure 1 illustrates the workflow of F-RLP, which benefits significantly from being trained with counterfactual data, in contrast to RLP. RLP, being trained solely on the user's factual data, is limited to predicting user behavior without offering any enhancements in terms of nutritional intake or preferences. Additionally, if a user receives dietary guidelines from a healthcare professional or advisor, there is no assurance that their historical data reflects adherence to these recommendations. Hence, RLP cannot guarantee compliance with such dietary instructions, as its operation is purely based on factual user logs.
Conversely, F-RLP incorporates a CFG component that utilizes expert advice to select the optimal choice, focusing on nutritional value and personal preferences, thus ensuring improvement. The CFG takes both the list of contextual options and the personal vector as inputs, generating counterfactual training data. This data is then used to retrain the text-to-text transformer, enhancing its ability to make recommendations that are not just based on past behavior but also aligned with expert dietary advice and improved nutritional goals.

\subsection{Query Stage}
As previously discussed, we utilize the T5 model as the foundational framework for our F-RLP system. After retraining the T5 model with counterfactual data, it's primed for querying. The training phase involved using a combination of the options list and the personal vector as inputs, with counterfactual data serving as the output. Consequently, to query the retrained model effectively, we require both a list of contextualized options and the personal vector at the query stage. This setup enables the retrained model to apply the insights gained from the counterfactual training phase, allowing it to select the most suitable option based on the comprehensive criteria it has learned.

\section{CFG Setup and Configuration}
\subsection{CFG Input}

The CFG process necessitates two primary inputs: the Personal Vector, derived from the individual's personal dataset, and the option list, generated by the CRE.
We engage in an N-of-1 intensive longitudinal study of personal data collected by an adult male, encompassing his food log and sensory data (including sleep patterns, physical activity levels, and heart rate) over a two-year period. Such comprehensive data collection is made possible through the use of food logging applications \cite{Rostami2020FL}, which gather a wide range of food-related information. This dataset encompasses every food item and ingredient consumed by the individual daily, alongside biometric data obtained from an Oura ring.
The Personal Vector itself is divided into two segments: one segment includes straightforward numerical data representing the average biometric readings over the last three days, while the other segment compiles a list of the individual's most favored ingredients based on consumption patterns observed over the last thirty days.

The second input for the CFG process is a list of contextual options, produced by the CRE. While this paper does not delve into the intricate implementation details of various CRE components, we simplify the approach by selecting twenty random food choices from the dataset referenced as \cite{marin2021recipe1m+} each time. These options are then prioritized based on the Personal Vector as well as any specified settings, ensuring that the recommendations are both relevant and personalized to the individual's preferences and context.

\begin{figure}
    \centering
    \includegraphics[width=0.4\textwidth]{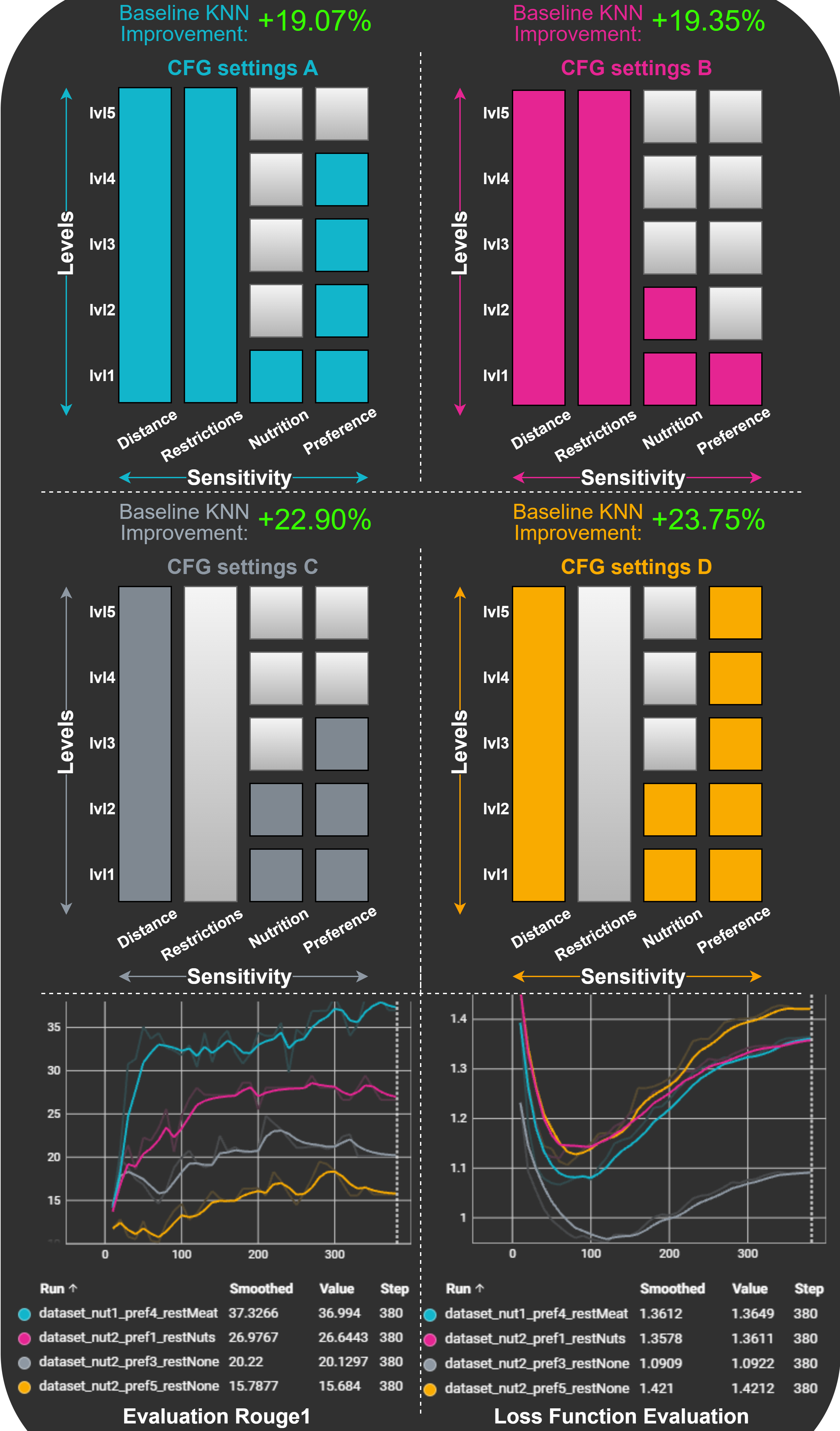}
    \caption{For each category, we utilize comparison metrics to assess the degree of improvement achieved by different setting classes relative to the baseline model without Counterfactual Generation (No CFG). The results indicate a positive enhancement across all categories. The horizontal axis of the charts represents the specific item selected to gauge sensitivity levels, while the vertical axis measures the extent of the sensitivity level itself.}
    \label{fig:eval}
    \vspace{-20pt}
\end{figure}

\begin{figure*}[t]
    \centering
    \includegraphics[width=\textwidth]{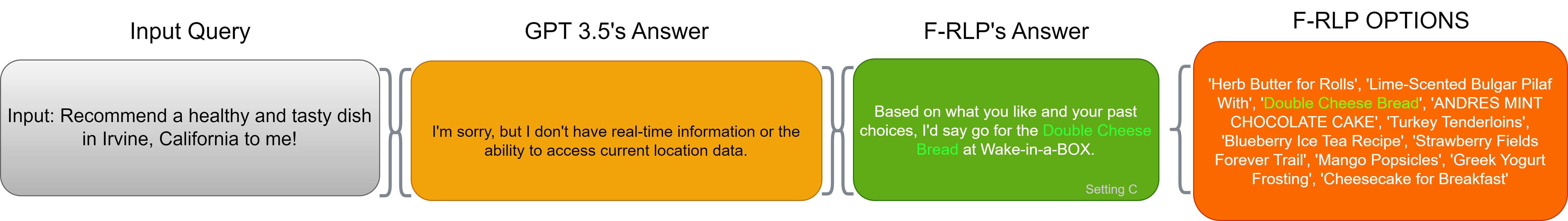}
    \caption{Showcasing F-RLP's meaningful and precise answer comparison with GPT 3.5 on our text food query.
}
    \label{fig:egg}
\end{figure*}
\subsection{CFG Settings}
The CFG process prioritizes four key factors when ranking the list of options, as follows:
\begin{itemize}
    \item Distance: This factor evaluates the significance of the proximity of the food choice (e.g., restaurant) to the user.
    \item Restrictions: This factor accounts for the presence of restricted ingredients, such as allergens or ingredients specifically advised against by a healthcare provider.
    \item Nutrition: This factor considers the nutritional content of the food in the decision-making process.
    \item Preference: This factor assesses the user's personal taste preferences in the final decision.
\end{itemize}
The CFG settings enable the assignment of different sensitivity levels to each of these factors, where a factor assigned a higher sensitivity level will exert a more substantial influence on the ranking of the options list. Except for the distance and restriction factors, which are binary, there are five sensitivity levels available for each factor.

In this experiment, the distance factor is always activated. This is because considering the proximity of food options is essential when generating the list in the CRE component. Given that we are working with a predetermined list of options for this study, we operate under the assumption that the distance factor has already been accounted for by providing a list of food choices that are within a reachable distance for the user.  

The restriction factor operates on a binary basis, indicating whether there is a specific list of ingredients that must be completely avoided or if no such restrictions exist. Figure 2 illustrates four sample setting profiles, along with the outcomes of each corresponding model trained under these conditions, presented on the left side.
In setting A, \textit{meat} was identified as a dietary restriction, whereas \textit{nuts} were the focus in setting B. The list of restricted items for setting A included "Pork," "Beef," "Ham," "Cow," "Lamb," "Chicken," "Steak," "Burger," "Hotdog," "Goat," "Turkey," "Bacon," "Sausage," and "Rib." For setting B, the restrictions encompassed "Nuts," "Seeds," "Pecans," "Almonds," and "Pistachios." These settings illustrate how specific dietary constraints are applied and managed within the CFG process to tailor the recommendation system to individual dietary needs and preferences.

\subsection{CFG Output}
The distance factor is pre-considered within the CRE component, ensuring that only food options within an accessible range to the user are included. Meanwhile, the restriction factor is applied by excluding any food items that contain the specified restricted ingredients. The nutrition and preference factors, unlike the binary distance and restriction factors, are assigned integer values ranging from zero to five to determine their priority levels.
The algorithm employed to rank the remaining options list based on these non-binary factors initially prioritizes the factor with the higher sensitivity, organizing the list accordingly. Subsequent selection from the list depends on the actual level value assigned to each factor. For instance, if a factor's level value is two, the list is halved, and the top portion is selected; if the value is three, the list is divided by three, and so forth. Following this initial sorting, the list is then reorganized according to the second factor using the same method, resulting in a fully sorted list.
This final list, produced by the CFG, serves as the basis for retraining the F-RLP model.

\section{Results}

The challenge of general food recommendation, hindered by an almost infinite number of classes, limited sample sizes, and unbalanced data distribution, renders most classification models ineffective. However, our proposed framework demonstrates significant promise over traditional approaches. This is substantiated by a comparative analysis against a KNN classifier using the same dataset. We evaluate the performance by measuring the deviation of the model's top recommendation from the CFG-sorted list and calculating the error rate in comparison to the baseline KNN model. Figure 2 offers crucial insights into the effectiveness of our methodology, highlighting its superior performance.
\begin{itemize}
    \item The sensitivity axis can be customized by each individual, leading to highly personalized results. This adjustment allows for a tailored experience that aligns closely with each user's unique preferences and needs.
    \item Contextual Relevance: By presenting a curated set of options to the LLM recommender with each query, we ensure that the context is integrated from the outset. This method acknowledges the significance of context, including location, in the selection process. 
    \item The model, retrained with counterfactual data, is designed to prioritize user satisfaction.
\end{itemize}
Figure 3 presents a sample query and response from F-RLP, illustrating its precision and relevance, a result of the effective integration between the option list and the LLM within our framework. It also contrasts this query's outcome with those from GPT 3.5 \cite{NEURIPS2020_1457c0d6}. 

\section{Conclusions}
In conclusion, while traditional food recommendation systems struggle with limitations of scale and data, the rise of LLMs introduces a promising solution. Yet, the general application of LLMs falls short in the specialized domain of food recommendations. Our F-RLP framework addressed this by tailoring LLM capabilities specifically for food, enhancing both accuracy and personalization. F-RLP represented a significant advancement in food recommendation technology, bridging the gap between generic algorithms and the specific needs of dietary guidance, and setting a new benchmark for precision in the field.


\bibliographystyle{IEEEtran}
\bibliography{IEEEabrv,ref}

\end{document}